\documentclass[letterpaper, 10 pt, conference]{ieeeconf}  

\IEEEoverridecommandlockouts                              

\usepackage{graphicx}
\usepackage{float}
\usepackage[caption=false]{subfig}

\usepackage{caption}
\usepackage{threeparttable}
\usepackage{cite}
\usepackage{url}
\makeatletter
\let\NAT@parse\undefined
\makeatother

\usepackage{multirow}
\usepackage{booktabs}
\usepackage{threeparttable}

\usepackage[breaklinks=true,colorlinks,bookmarks=false]{hyperref}
\usepackage{algorithm}
\usepackage{algorithmic}
\usepackage{soul}
\usepackage{amsmath}

\title{\LARGE \bf
PCB-RandNet: Rethinking Random Sampling for LiDAR Semantic Segmentation in Autonomous Driving Scene 
}

\author{Xian-Feng Han$^{\ast}$, Huixian Cheng$^{\ast\dag}$, Hang Jiang, Dehong He, Guoqiang Xiao 
\thanks{All authors are with College of Computer and Information Science, Southwest University, Chongqing, China. This work was supported by the National Natural Science Foundation of China (No. 62002299), and the Natural Science Foundation of Chongqing, China (No. cstc2020jcyj-msxmX0126), and the Fundamental Research Funds for the Central Universities (No. SWU120005)}%
\thanks{$^{\ast}$ Equal Contribution $^{\dag}$ Now working at Seyond (formerly Innovusion)}
}

\begin{document}

\maketitle
\thispagestyle{empty}
\pagestyle{empty}

\begin{abstract}

Fast and efficient semantic segmentation of large-scale LiDAR point clouds is a fundamental problem in autonomous driving. To achieve this goal, the existing point-based methods mainly choose to adopt Random Sampling strategy to process large-scale point clouds. However, our quantative and qualitative studies have found that Random Sampling may be less suitable for the autonomous driving scenario, since the LiDAR points follow an uneven or even long-tailed distribution across the space, which prevents the model from capturing sufficient information from points in different distance ranges and reduces the model's learning capability. To alleviate this problem, we propose a new \textbf{P}olar \textbf{C}ylinder \textbf{B}alanced \textbf{Rand}om Sampling method that enables the downsampled point clouds to maintain a more balanced distribution and improve the segmentation performance under different spatial distributions. In addition, a sampling consistency loss is introduced to further improve the segmentation performance and reduce the model's variance under different sampling methods. Extensive experiments confirm that our approach produces excellent performance on both SemanticKITTI and SemanticPOSS benchmarks, achieving a $2.8\%$ and $4.0\%$ improvement, respectively. The source code is available at \href{https://github.com/huixiancheng/PCB-RandNet}{PCB-RandNet}.

\end{abstract}

\section{INTRODUCTION}
Recently, with the rapid development of 3D computer vision, great advances have been achieved in autonomous driving ~\cite{geiger2012we, feng2020deep}, which highly depends on accurate, robust, reliable, and real-time 3D semantic perception and understanding of surroundings. For all sensors equipped in autonomous vehicles, LiDAR sensors play an important role. Therefore, LiDAR-based semantic segmentation task, producing point-wise semantic labels, has been attracting more and more attention. Currently, the existing approaches can be summarized into three categories:  Voxel-based~\cite{choy20194d}, Projection-based~\cite{wu2018squeezeseg, milioto2019rangenet++, li2021multi}, and Point-based methods~\cite{qi2017pointnet, qi2017pointnet++}.

Voxel-based and Projection-based methods inevitably introduce information loss in transforming 3D point clouds into structured voxel and 2D spatial representations.

\begin{figure}[t]
	\subfloat[]{
	\label{fig_a}
	\includegraphics[width=0.5\linewidth, height=2.8cm]{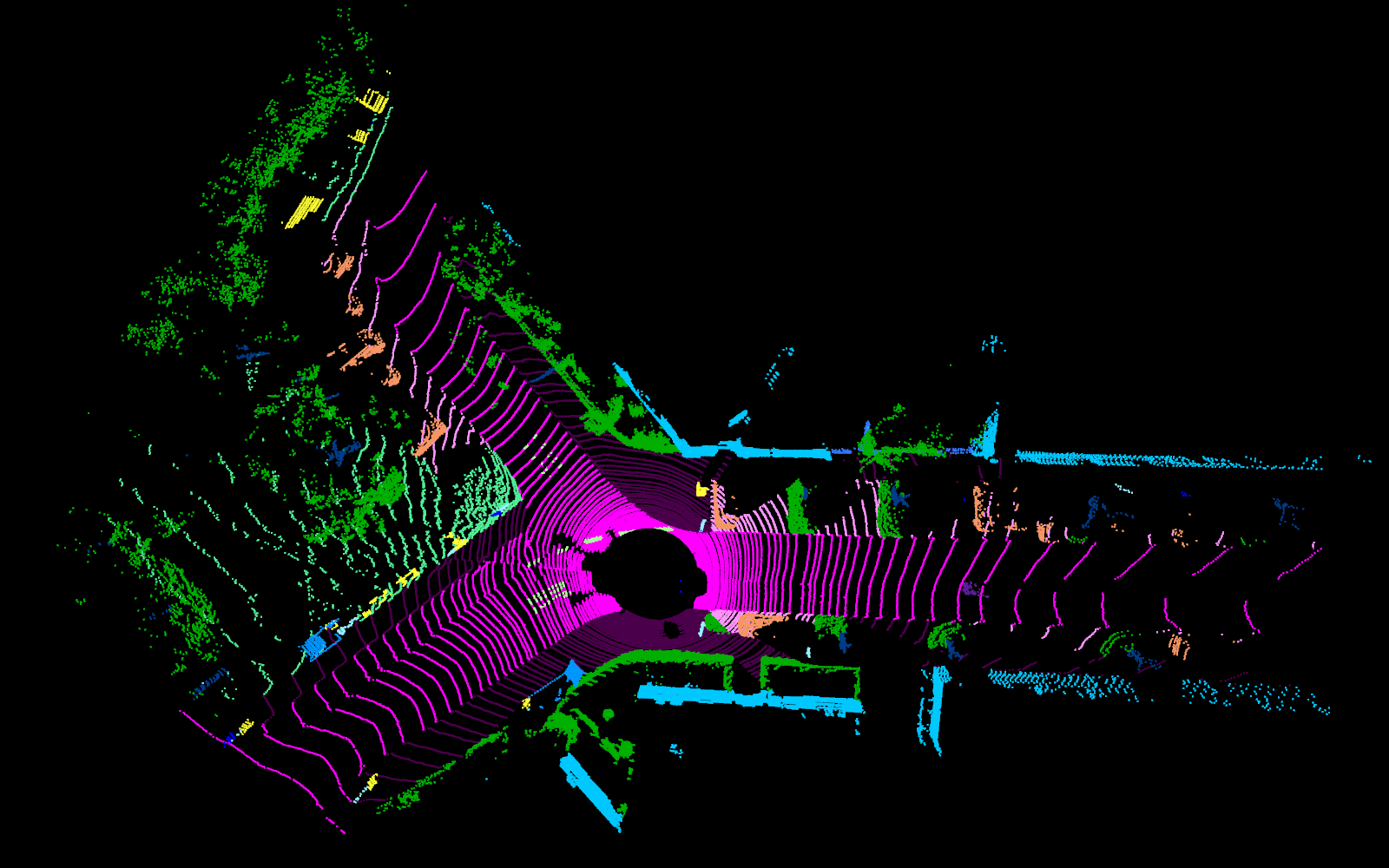}}
\subfloat[]{%
	\label{fig_b}
	\includegraphics[width=0.5\linewidth, height=2.8cm]{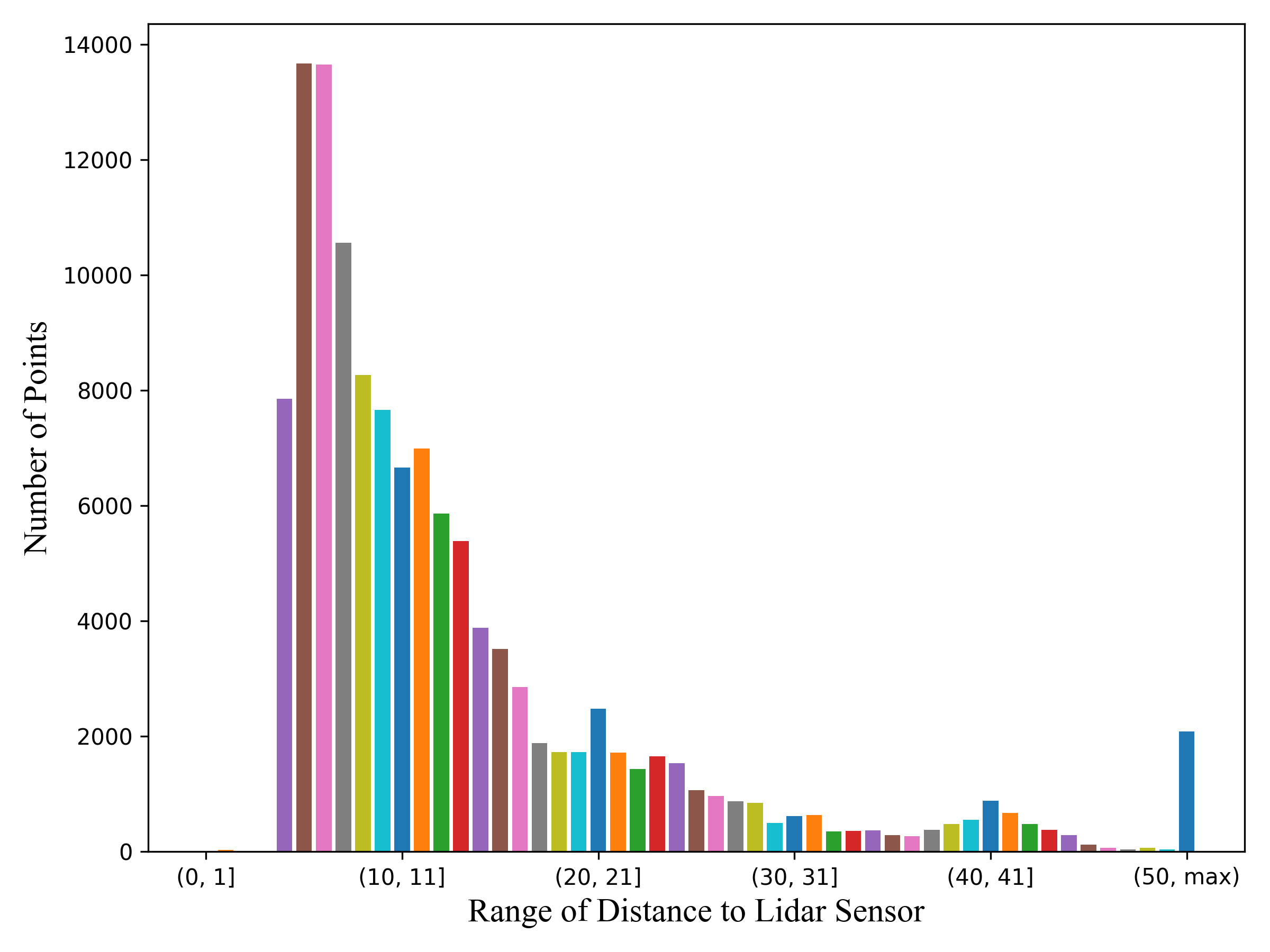}}
\\\\[-2ex]
\subfloat[]{%
	\label{fig_c}
	\includegraphics[width=0.5\linewidth, height=2.8cm]{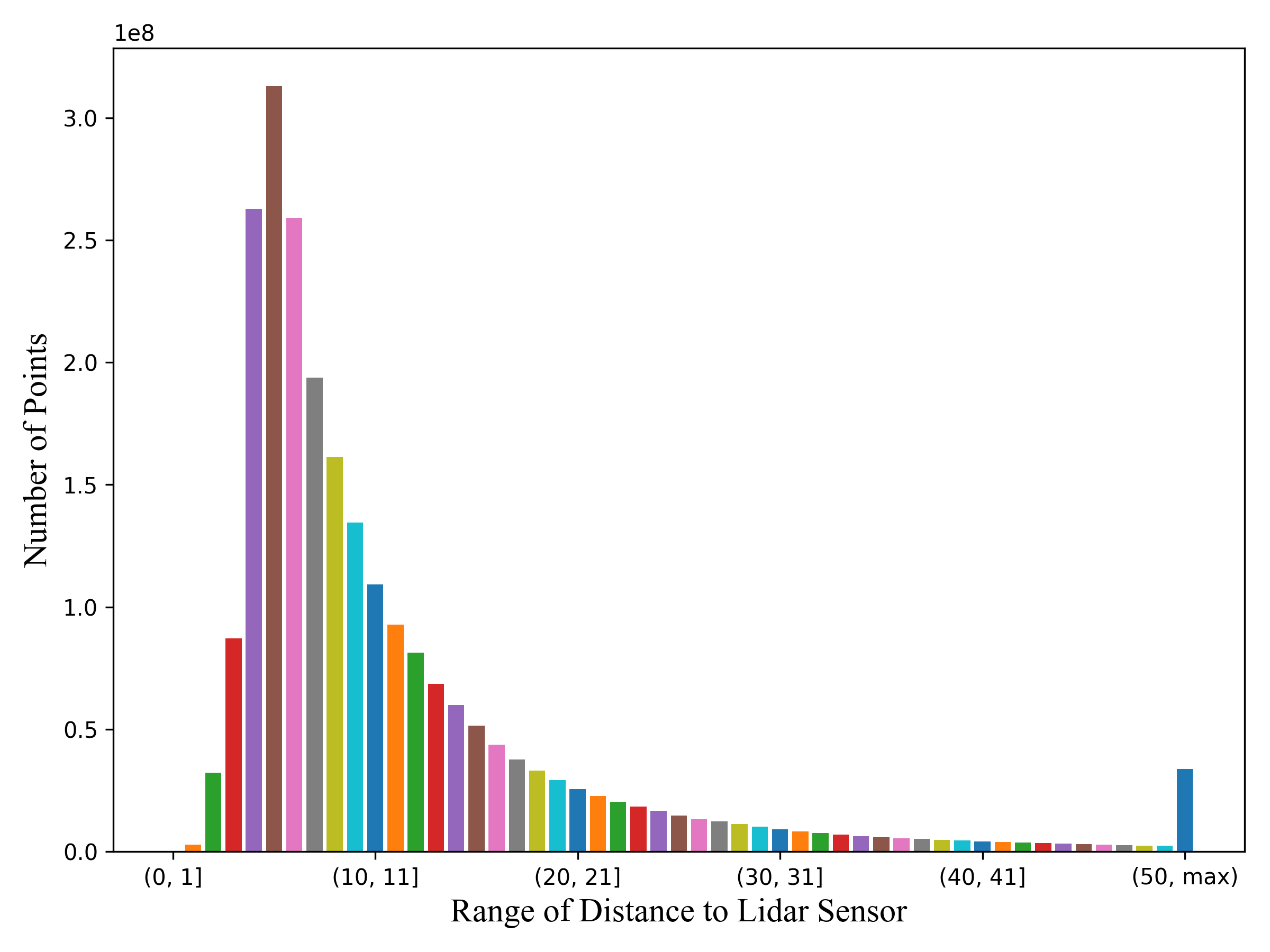}}
\subfloat[]{%
	\label{fig_d}
	\includegraphics[width=0.5\linewidth, height=2.8cm]{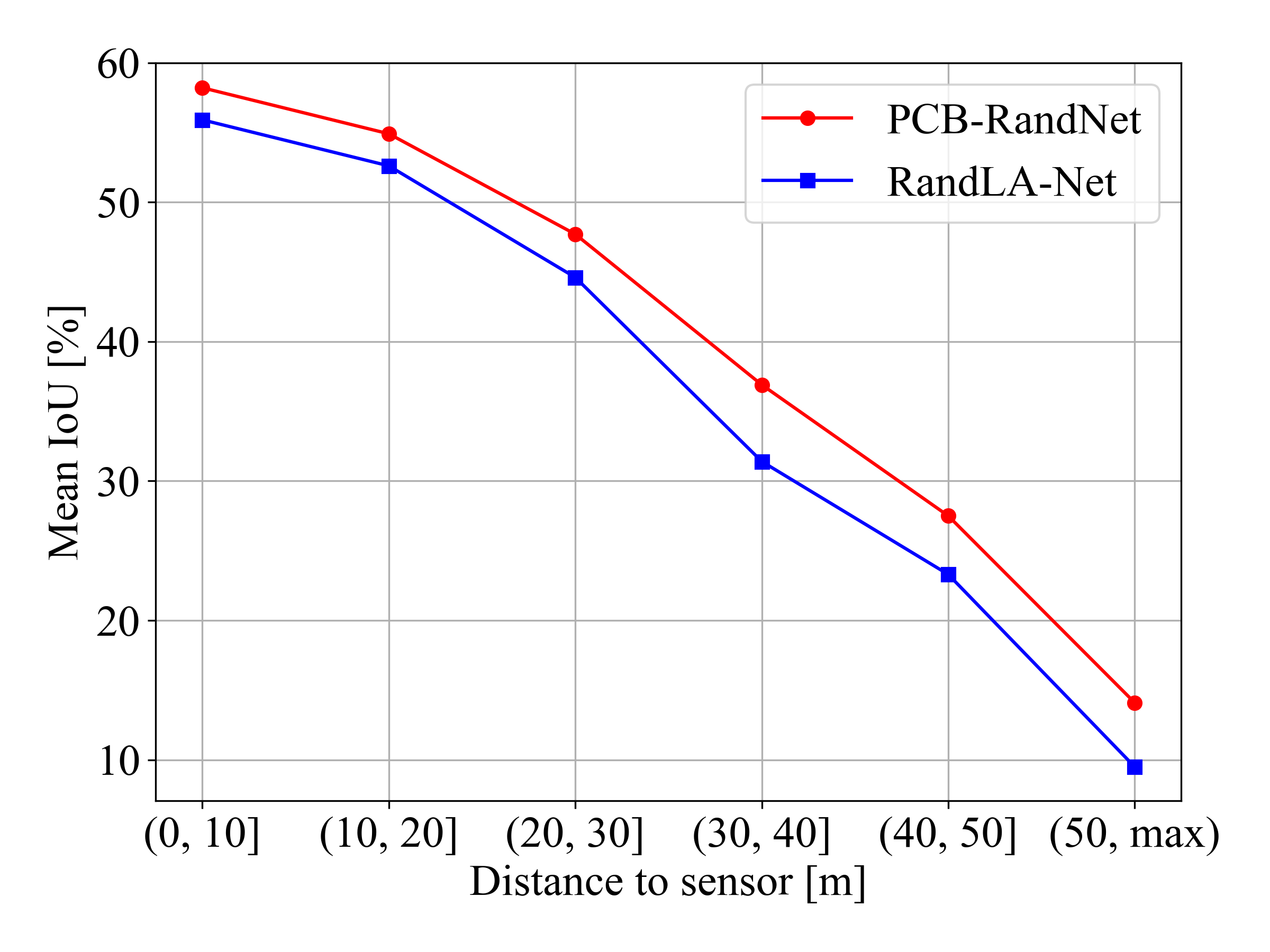}}
\caption{\textbf{(a)}: Visualization of one frame of LiDAR scan on the SemanticKITTI dataset. \textbf{(b)}:Visualization of the corresponding point cloud scan distance distribution. \textbf{(c)}: Visualization of the distance distribution of all training data. \textbf{(d)}: Segmentation performance of the baseline method and our method at different distances.}
\label{fig:first}
\vspace{-5mm}
\end{figure}

Point-based methods can directly deal with raw disordered, unstructured 3D point clouds without any information loss. Although these methods have produced encouraging results in terms of object classification and part segmentation, most are only suitable for small-scale point clouds or indoor scenes, and cannot be directly extended to large-scale cases. This is mainly because the usage of the farthest point sampling (FPS) operation causes much higher computational complexity and memory consumption ~\cite{qi2017pointnet++, li2018pointcnn, wu2019pointconv, yan2020pointasnl}. 
To address this problem, RandLA-Net~\cite{hu2020randla} proposes a new efficient and lightweight neural architecture by combining simple \textbf{Random Sampling (RS)} and local feature aggregation modules, which achieves high efficiency and state-of-the-art performance on multiple benchmarks.

However, when revisiting the random sampling method used in RandLA-Net for the task of outdoor LiDAR semantic segmentation, we get the following findings. Taking a randomly selected frame (Seq:00 ID:0000) of point cloud data (shown in Figure~\ref{fig_a}) from the SemanticKITTI dataset as an example, we quantitatively visualize its density (or the number of points) distribution with respect to distance between points and LiDAR sensor. And it clearly follows a long-tail distribution. That is, the closer the distance to sensor is, the point clouds are much more dense. 

Here is one question: \textbf{Is random sampling still applicable to such a specific autonomous driving scenario?} Obviously, the answer is \textbf{No}. Because Random Sampling selects $M$ points evenly from the initial $N$ points, where each point has the same probability of being selected ~\cite{hu2020randla}. However, we have found that points at close distances have a high possibility of being retained, while points at medium and long distances tend to be discarded, which results in reduction of model's capability of learning sufficient representation from points in different distance ranges. As shown in Figure~\ref{fig_d}, it can be reported that the segmentation performance of the model decreases sharply as the distance to the sensor increases.

To address this problem, we propose a novel sampling method, termed \textbf{Polar Cylinder Balanced Random Sampling} (PCB-RS). Specifically, we first divide the input point cloud into different cylindrical blocks in the polar BEV view, and then perform the balanced random sampling within each block. The downsampled point cloud can maintain a more balanced distribution, which guides the network to sufficiently learn point features at different distances, and improve the segmentation performance within different distance ranges (as shown in Figure~\ref{fig_d}). In addition, we present a \textbf{Sampling Consistency Loss} (SCL) function to reduce the model variation under different sampling methods. 

We evaluate our proposed approach on two large-scale outdoor autonomous driving scenario datasets, namely SemanticKITTI and SemanticPOSS. Experimental results show that our method improves the baseline RandLANet by 2.8 mIoU on the SemanticKITTI dataset and by 4 mIoU on the SemanticPOSS dataset.

The main contributions of this work can be summarized as follows:
\begin{itemize}
	\item We propose a novel Polar Cylinder Balanced Random Sampling method to downsample the point clouds, which contributes to guiding the segmentation model to better learn the point distribution characteristics in autonomous driving scenario.
	
	\item  We propose a sampling consistency loss function to redcue the model's variance under different sampling approaches.
	
	\item Experimental results on SemanticKITTI and SemanticPOSS demonstrate the superior performance of our proposed method.
\end{itemize}

\begin{figure*}[ht]
	\centering
	\includegraphics[width=\textwidth]{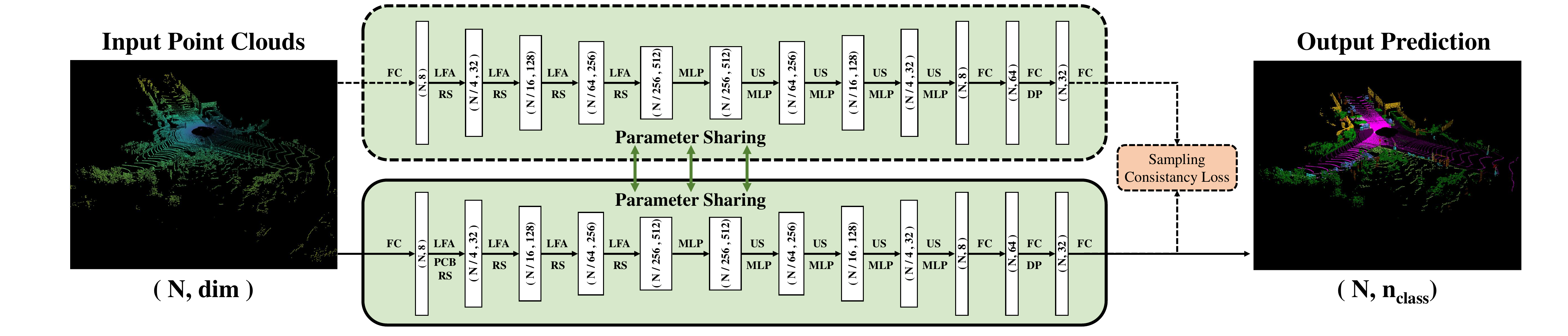}
	\caption{\textbf{Overall pipeline of PCB-RandNet}. Our network structure remains the same as RandLA-Net, with \textbf{only changes to the first sampling method}.
		Two different class prediction distributions $P_1$ and $P_2$ can be obtained by forwarding the input point cloud twice by models with different sampling methods but shared parameters. 
		Then, the sampling consistency loss between the two class prediction probabilities is calculated and used to supervise the model.
		$(N,D)$ represents the number of points and feature dimension respectively. RS: Random Sampling, PCB-RS: Polar Cylinder Balanced Random Sampling, FC: Fully Connected layer, LFA: Local Feature Aggregation, MLP: shared Multi-Layer Perceptron, US: Up-sampling, DP: Dropout.
        (\textit{Please note that as shown by the dashed line, the upper branch and sampling consistency loss is only used during training and discarded during inference, i.e., it is disposable. Therefore, it does not introduce any computational burden during inference.})
	}
	\label{fig:framework}
	\vspace{-4mm}
\end{figure*}

\begin{figure*}[ht]
	\centering
	\includegraphics[width=\linewidth]{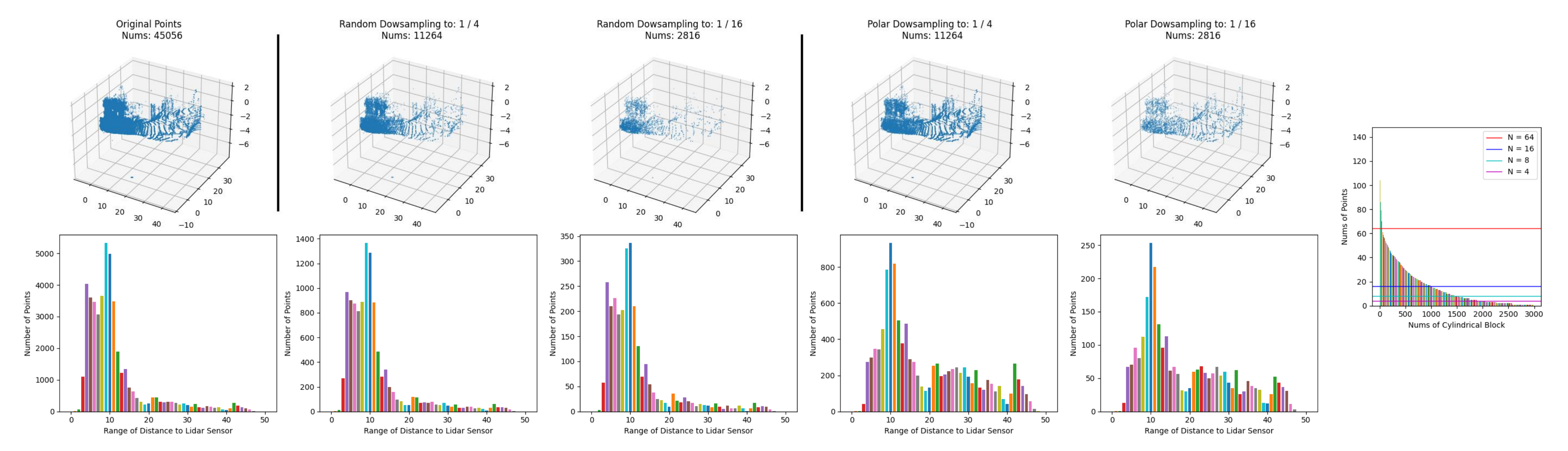}
	\caption{Comparison of qualitative visualization and quantitative statistical analysis between RS and PCB-RS. The upper part shows the visualization of the point cloud, and the lower part shows the distribution of that point cloud at different distance ranges. The subplot on the right shows the distribution of the number of points within the cylindrical block when using the PCB-RS method. \textit{Best viewed in color and zoomed in for more detail.}}
	\label{fig:sample}
	\vspace{-6mm}
\end{figure*}

\section{Related work}
\noindent\textbf{Voxel-based Methods} first discretize point clouds into 3D voxel representations and then predict the semantic labels for these voxels using 3D CNN frameworks. 
SPVNAS~\cite{tang2020searching} utilizes neural structure search (NAS) to further improve the performance of the network. AF2-S3Net~\cite{cheng20212} introduces two attention blocks to learn local and global contexts and highlight detailed information. 
Cylinder3D~\cite{zhu2021cylindrical} and SphereFormer~\cite{lai2023spherical} explore more adequate voxel division methods.

\noindent\textbf{Projection-based Methods} typically perform spherical projection to convert LiDAR point clouds into regular 2D images and employ 2D CNNs segmentation methods.
SqueezeSeg~\cite{wu2018squeezeseg, wu2019squeezesegv2, xu2020squeezesegv3} series methods and RangeNet++~\cite{milioto2019rangenet++} are seminal work for such methods. Subsequent works~\cite{cortinhal2020salsanext, razani2021lite, zhao2021fidnet, cheng2022cenet, ando2023rangevit, kong2023rethinking} further explored the network structure.


\noindent\textbf{Point-based Methods} take unordered points as input and predict point-wise semantic labels. PointNet~\cite{qi2017pointnet} and PointNet++~\cite{qi2017pointnet++} are pioneering studies that utilize shared MLPs to learn the point-wise properties, which greatly boosts the development of point-based networks. KPConv~\cite{thomas2019kpconv} develops deformable convolution, which can use a flexible approach to learn local representations. 
RandLA-Net~\cite{hu2020randla} uses random sampling method to processing large-scale point clouds and uses local feature aggregation to reduce the information loss caused by random operations. BAF-LAC~\cite{shuai2021backward} proposes a backward attention fusion encoder-decoder that learns semantic features and a local aggregation classifier. BAAF~\cite{qiu2021semantic} obtains more accurate semantic segmentation by using bilateral structures and adaptive fusion methods that take full advantage of the geometric and semantic features of the points. 

\section{Methodology}
\subsection{Overview}
The overall structure of our PCB-RandNet model is illustrated in Figure~\ref{fig:framework}. We will detailedly describe the Baseline Network, Polar Cylinder Balanced Random Sampling and Sampling Consistency Loss in the following subsections.

\subsection{Baseline Network}
In order to verify the effectiveness of our proposed sampling strategy, we choose to use RandLA-Net~\cite{hu2020randla} as the baseline network model based on the following two considerations: 1) since our PCB-RS sampling method can be considered as an extension of the RS method, and RandLA-Net is the first point-based method that has achieved competitive performance on the SemanticKITTI dataset using random sampling method, it is reasonable to use RandLA-Net as baseline. 2) In ~\cite{hu2021learning}, Hu et al. conducted ablation experiments on the SemanticKITTI dataset to evaluate the effects of different sampling strategies (FPS, IDIS, PDS, GS, CRS, PGS) had on segmentation performance. The experimental results show that RS outperforms other sampling methods, which allows us to focus on the comparison with RS method only. It should be specially noted that our approach actually can be well embedded in other networks, and bring significant improvement as shown in Section~\ref{sec:exper} and Section~\ref{sec:abla}.  

\begin{figure}[t]
\begin{algorithm}[H]
	\caption{Simple example algorithm of RS and PCB-RS}
	\label{alg:sampling}
	\textbf{Input}: Points with coordinates: $N \times 3$\\
	\textbf{Output}: Sampling points: $M \times 3$ \\
	
	\textbf{Random Sampling}: 
	\begin{algorithmic}[1] 
		\STATE Random shuffle points.
		\STATE Slice select the top $M$ points.
		\RETURN Sampling points $M \times 3$
	\end{algorithmic}
	\texttt{\\}
	\textbf{PCB Random Sampling}:
	\\
	\textbf{New Param}: Resolution of the polar cylindrical blocks representation: $R \times P \times Z$ 
	\begin{algorithmic}[1] 
		\STATE Convert Cartesian to Polar coordinates.
		\STATE Dividing the point cloud into different cylindrical blocks based on the resolution parameter size $R \times P \times Z$ .
		\texttt{\\}
		\textit{// \textcolor{green}{Here will get the number of blocks $K$.}}
		\STATE Based on target sampling points $M$ and total number of blocks $K$, calculate the number of points to be individually downsampled for each cylindrical block. \textit{(Keep number of points closer to each other.)}
		\texttt{\\}
		\textit{// \textcolor{green}{Here will get a set of sample number list $S_{n}$: $K \times 1$.}}
		\STATE {$Sampling \gets [None]$}
		\FOR { $i \gets 1$ to $K$}
		\STATE {Get all sub points $N_{i}$ located in this block.}
		\STATE {Get the number of points $S_{n}[i]$ to be sampled within the block.}
		\STATE Random shuffle sub points $N_{i}$.
		\STATE Slice select the top $S_{n}[i]$ points.
		\STATE Splice this subsampling point. 
		\texttt{\\}
		\textit{// \textcolor{green}{e.g. $Sampling.append()$}}
		\ENDFOR
		\STATE Random shuffle Sampling points $Sampling$.
		\RETURN Sampling points $M \times 3$
	\end{algorithmic}
\end{algorithm}
\vspace{-8mm}
\end{figure}

\subsection{Polar Cylinder Balanced Random Sampling}
\label{sec:pcbrs}
From previous discussion, we clearly see that the LiDAR point clouds from autonomous driving scenes have varying density, i.e. the nearby regions have much larger density than farther areas. And the simple RS strategy may not be an ideal choice for this situation, since the randomly downsampled points usually follow the same probability distribution as that of original point cloud, which means that the RS cannot deal with the unbalanced distribution problem. Figure~\ref{fig:sample} provides visualizations and statistical analysis of a case. It can be clearly and intuitively stated that the point clouds obtained by downsampling  $1 / 4$ and $1 / 16$ follow the same distance distribution as the original point cloud, which leads to most of the sampled points aggregating in the close-range regions. And the network with random sampling is likely to \textit{``over-fit''} the point cloud at close distances and \textit{``under-fit''} the points at medium and long distances.

Therefore, in order to address the problem above, we propose a new sampling method named \textbf{Polar Cylinder Balanced Random Sampling} (PCB-RS) inspired by the basic concepts of related works~\cite{zhang2020polarnet,zhou2020cylinder3d,zhu2021cylindrical}. The core idea behind our PCB-RS is that the points in the downsampled point cloud should be distributed as uniform as possible over different distance ranges, so that the segmentation model is able to learn adequate information from all distances. 

The Algorithm~\ref{alg:sampling} describes the detailed process of our PCB-RS, and comparison with RS method. It can be obviously reported that RS simply shuffles the input point cloud with $N$ points, and then selects the top $M$ points as the target point cloud to perform sampling. In contrast, PCB-RS first converts the Cartesian coordinates into the Polar coordinate system, which transforms the point coordinate ($x, y, z$) to ($\rho, \theta, z$), where the radius $\rho$ (distance to the origin of the x-y axis) and the azimuth $\theta$ (angle from the x-axis to the y-axis) are calculated. Based on this constructed Polar Coordinate System, we divide the input into different cylindrical blocks, and the farther the distance is, the larger the cells are. The right part of Figure~\ref{fig:sample} visualizes the cylindrical blocks and the number of points in each block for an example point cloud, where a long-tail distribution is observed.

Based on the number of target sampling points $M$ and the number of cylindrical blocks $K$, we calculate the number of points ($S_{n}$: $K \times 1$) in each cylindrical block. \textbf{During downsampling, we attempt to balance the number of points per block, making them as the same as possible.} Then we perform the sampling operations similar to RS in each cylindrical block. To be specific, we shuffle the points within each block, and select the top $S_{n}[i]$ points to yield a sub point cloud. After combining all the resulted sub point clouds, we perform shuffle operation once again to obtain the sampled point cloud.

Through the usage of the PCB-RS sampling method, on one hand, we are able to guarantee that the distribution of the sampled points over different distance ranges can be as uniform as possible. From the visualization in Figure~\ref{fig:sample}, we find that our PCB-RS can change the point distribution with respect to distance to sensor by retaining the points at medium and long distances. 

On the other hand, in terms of computation complexity, PCB-RS have a low computation cost since it can be considered as a collection of $K$ sub-random sampling processes. Figure~\ref{fig:framework} presents the overall architecture of our segmentation network, \textbf{where we only use PCB-RS method in the first downsampling stage}, and other subsequent sampling layers still adopt RS approach. \textbf{The reasons are that:} (1) the distribution of point clouds processed by this PCB-RS layer have been sufficiently balanced at different distances. (2) In the subsequent stages, this balance distribution can be maintained well even if they still use RS strategy for downsampling. (3) This sampling configurations makes our network efficient.

\subsection{Loss and Sampling Consistency Loss}
LiDAR semantic segmentation datasets often suffer from class imbalance problem. For example, in the SemanticKITTI dataset, the proportions of roads and sidewalks are hundreds of times larger than that of people and motorcycles. To handle this issue, we choose to use weighted cross-entropy loss function to better optimize these rare categories, which is formulated as,
\vspace{-1ex}
\begin{align}
\mathcal{L}_{wce}=& -\sum_{c=1}^{C}\omega _{c} \cdot y_{c}log(\hat{y}_{c}),
\vspace{-1ex}
\end{align}
where $C$ is the total number of classes in the dataset, $\omega _{c}$ is the calculated weight of the $c$th class, $y_{c}$ and $\hat{y}_{c}$ denote the ground truth and prediction probability, respectively.


In addition, inspired by studies~\cite{wu2021r, li2022cpgnet}, we design a simple \textbf{Sampling Consistency Loss} to guide the model to learn the variability of different sampling methods, and enhance the model's representation power by learning more diverse feature representations. Meanwhile it constrains the model to output a probability distribution as consistent as possible under different sampling methods. 
The formula for sampling consistency loss is defined as,
\vspace{-1ex}
\begin{equation}
\mathcal{L}_{scl}=\sum_{c=1}^{C}|\hat{y}_c^{PCB\text{-}RS} - \hat{y}_c^{RS}|,
\vspace{-1ex}
\end{equation}
where $\hat{y}_c^{PCB\text{-}RS}$ and $\hat{y}_c^{RS}$ are the predicted probabilities obtained by using PCB-RS and RS, respectively. The total loss $\mathcal{L}_{total}$ is the weighed sum of these two loss functions, formulated as, 
\vspace{-1ex}
\begin{equation}
\mathcal{L}_{total}=\mathcal{L}_{wce} + \alpha\mathcal{L}_{scl}
\vspace{-1ex}
\end{equation}

However, such a loss formula leads to a more tedious optimal weighted hyperparameter search process. In order to solve this problem, we introduce the uncertainty weighting method proposed by~\cite{kendall2018multi}, which automatically adjusts the optimal ratio between different loss terms by introducing two learnable parameters $\sigma_1$ and $\sigma_2$. And we also add two additional logarithmic regular terms to maintain stability. 
The total loss can be written as follows,

\begin{equation}
	\label{eq:learn}
	\begin{aligned}
		\mathcal{L}_{total} = \frac{1}{\sigma_1^2}\mathcal{L}_{wce}+\frac{1}{\sigma_2^2}\mathcal{L}_{scl}\\
		+log (1 + \sigma_1 ) + log (1 + \sigma_2 ),
	\end{aligned}
\end{equation}

\section{Experiments} 
\label{sec:exper}
To verify the effectiveness of our approach, we evaluate it on two benchmarks, SemanticKITTI~\cite{behley2019semantickitti} and SemanticPOSS~\cite{pan2020semanticposs}.




\subsection{Implementation Details.}
We perform all experiments on a single RTX 3090 GPU. During training, we use the same experimental setup as the RandLA-Net. The network is trained for 100 epochs using Adam as the optimizer. The initial learning rate is set to 0.01 and reduced by 5\% after each epoch. 

For both datasets, the resolution of the polar cylinder representation used in PCB-RS is set to 64 $\times$ 64 $\times$ 16, where these three dimensions correspond to radius, angle, and height, respectively. For SemanticKITTI, we keep the point cloud in the range of $[distance: 3 \sim max(\cdot) m, z:-3.0 \sim 1.5m]$. For SemanticPOSS, the point cloud in kept in the range of $[distance: 3 \sim 80m, z:-3.0 \sim 3m]$. 

For a fair comparison, we re-implement RandLA-Net~\cite{hu2020randla}, BAF-LAC~\cite{shuai2021backward} and BAAF~\cite{qiu2021semantic} based on the PyTorch framework.

\subsection{Semantic Segmentation Results}
\label{sec:result}

\begin{table*}[htb]
	\vspace{-5mm}
	\centering
	\caption{Evaluation results on SemanticKITTI validation set at different distance ranges. }
	\label{tab:result_kitti}
	\resizebox{\textwidth}{!}{%
		\begin{tabular}{r|ccccccccccccccccccccc}
			\toprule[1.5pt]
			& Method & \rotatebox{90}{\textbf{mIoU(\%)}} & \rotatebox{90}{car} & \rotatebox{90}{bicycle} & \rotatebox{90}{motorcycle} & \rotatebox{90}{truck} & \rotatebox{90}{other-vehicle} & \rotatebox{90}{person} & \rotatebox{90}{bicyclist} & \rotatebox{90}{motorcyclist} & \rotatebox{90}{road} & \rotatebox{90}{parking} & \rotatebox{90}{sidewalk} & \rotatebox{90}{other-ground} & \rotatebox{90}{building} & \rotatebox{90}{fence} & \rotatebox{90}{vegetation} & \rotatebox{90}{trunk} & \rotatebox{90}{terrain} & \rotatebox{90}{pole} & \rotatebox{90}{traffic-sign} \\
			\toprule[1.5pt]
			\multirow{4}{*}{Range From 0 m to 10 m} & RS & 55.9& 96.1 & 13.3 & 32.6 & 61.1 & 45.5 & 59.6 & 76.0 & 0.0 & 94.2 & 43.9 & 81.8 & 0.0 & 86.4 & 59.1 & 85.6 & 71.8 & 70.1 & 62.8 & 22.4  \\
			\cmidrule{2-22}
			& PCB-RS & \textbf{58.9}& 96.4 & 11.2 & 51.0 & 75.8 & 53.3 & 65.8 & 84.1 & 0.0 & 93.5 & 42.6 & 83.1 & 0.0 & 87.9 & 61.1 & 86.6 & 67.7 & 74.9 & 58.6 & 25.5 \\
			\cmidrule{2-22}
			& PCB-RS+SCL & 58.2& 95.7 & 5.8 & 41.4 & 86.5 & 40.2 & 52.0 & 83.8 & 0.0 & 94.7 & 42.7 & 84.7 & 0.0 & 85.8 & 64.9 & 89.1 & 73.8 & 76.5 & 67.3 & 20.4 \\
			\toprule[1.5pt]		
			\multirow{4}{*}{Range From 10 m to 20 m} & RS & 52.6& 93.3 & 14.2 & 16.7 & 65.2 & 28.4 & 51.5 & 66.5 & 0.0 & 91.0 & 43.9 & 70.4 & 1.6 & 91.2 & 33.1 & 84.1 & 68.0 & 71.5 & 54.6 & 54.9   \\
			\cmidrule{2-22}
			& PCB-RS & 51.4& 90.6 & 15.8 & 24.2 & 25.1 & 32.9 & 49.6 & 78.8 & 0.0 & 91.8 & 45.6 & 72.4 & 0.0 & 91.6 & 33.4 & 83.0 & 64.1 & 69.5 & 54.3 & 54.0 \\
			\cmidrule{2-22}
			& PCB-RS+SCL & \textbf{54.9}& 94.2 & 10.1 & 18.1 & 58.2 & 36.3 & 51.1 & 82.2 & 0.0 & 92.9 & 47.3 & 74.4 & 0.1 & 92.1 & 40.6 & 85.9 & 70.4 & 71.9 & 59.2 & 57.7 \\
			\toprule[1.5pt]	
			\multirow{4}{*}{Range From 20 m to 30 m} & RS & 44.6& 82.4 & 2.0 & 4.2 & 76.9 & 18.6 & 29.8 & 47.9 & 0.0 & 82.4 & 29.9 & 54.6 & 1.3 & 86.9 & 20.5 & 82.0 & 57.6 & 72.7 & 46.6 & 51.3 \\
			\cmidrule{2-22}
			& PCB-RS &  45.5& 84.2 & 7.6 & 13.5 & 70.5 & 24.6 & 37.5 & 41.0 & 0.0 & 84.4 & 32.4 & 56.6 & 0.0 & 87.8 & 21.3 & 80.5 & 54.6 & 70.5 & 46.5 & 50.7 \\
			\cmidrule{2-22}
			& PCB-RS+SCL & \textbf{47.7}& 88.0 & 2.7 & 12.0 & 74.2 & 21.6 & 38.0 & 46.8 & 0.0 & 86.0 & 33.4 & 58.4 & 0.5 & 89.2 & 28.6 & 85.7 & 63.6 & 72.2 & 49.2 & 55.6 \\
			\toprule[1.5pt]	
			\multirow{4}{*}{Range From 30 to 40 m} & RS & 31.4& 58.2 & 1.6 & 0.6 & 22.1 & 16.5 & 14.1 & 8.8 & 0.0 & 71.4 & 16.1 & 38.3 & 8.6 & 81.3 & 11.4 & 78.6 & 44.5 & 63.9 & 36.6 & 23.3 \\
			\cmidrule{2-22}
			& PCB-RS & 33.3& 64.8 & 1.5 & 0.0 & 21.8 & 20.9 & 20.4 & 16.7 & 0.0 & 74.1 & 17.1 & 41.4 & 2.2 & 83.4 & 14.4 & 80.6 & 44.1 & 65.3 & 38.3 & 26.0 \\
			\cmidrule{2-22}
			& PCB-RS+SCL & \textbf{36.9}& 72.5 & 1.2 & 1.5 & 37.0 & 16.6 & 19.0 & 31.1 & 0.0 & 77.1 & 15.8 & 42.0 & 7.5 & 85.6 & 17.9 & 85.5 & 52.4 & 68.4 & 41.3 & 28.4 \\
			\toprule[1.5pt]	
			\multirow{4}{*}{Range From 40 to 50 m} & RS & 23.3& 38.8 & 0.0 & 0.0 & 4.2 & 15.9 & 2.5 & 1.1 & 0.0 & 61.3 & 9.0 & 25.8 & 4.2 & 77.4 & 6.8 & 77.3 & 32.0 & 52.7 & 23.5 & 11.0 \\
			\cmidrule{2-22}
			& PCB-RS & 25.5& 45.9 & 0.1 & 0.0 & 12.4 & 15.3 & 2.7 & 2.0 & 0.0 & 62.2 & 9.1 & 30.1 & 0.2 & 80.9 & 9.2 & 80.2 & 36.0 & 57.6 & 26.4 & 13.8 \\
			\cmidrule{2-22}
			& PCB-RS+SCL  & \textbf{27.5}& 52.0 & 0.1 & 0.0 & 13.5 & 15.7 & 4.5 & 2.0 & 0.0 & 66.3 & 5.2 & 29.0 & 5.4 & 83.4 & 11.3 & 84.9 & 40.5 & 60.3 & 31.6 & 16.6 \\
			\toprule[1.5pt]	
			\multirow{4}{*}{Range Above 50 m} & RS & 9.5& 13.6 & 0.0 & 0.0 & 0.0 & 5.6 & 0.0 & 0.0 & 0.0 & 18.8 & 2.6 & 9.9 & 0.0 & 16.3 & 2.0 & 68.5 & 3.5 & 23.7 & 16.4 & 0.0  \\
			\cmidrule{2-22}
			& PCB-RS & 11.4& 15.7 & 0.0 & 0.0 & 0.0 & 7.3 & 0.0 & 0.0 & 0.0 & 21.0 & 0.7 & 11.6 & 0.0 & 18.1 & 1.2 & 78.1 & 6.1 & 34.8 & 22.2 & 0.0 \\
			\cmidrule{2-22}
			& PCB-RS+SCL & \textbf{14.1}& 27.2 & 0.0 & 0.0 & 0.0 & 0.7 & 0.0 & 0.0 & 0.0 & 26.1 & 0.4 & 7.7 & 0.0 & 22.2 & 6.5 & 88.7 & 15.1 & 40.5 & 32.0 & 0.0 \\
			\toprule[1.5pt]	
			\multirow{4}{*}{Total Range} & RS & 53.7& 93.7 & 12.4 & 27.8 & 61.6 & 39.8 & 51.1 & 69.2 & 0.0 & 92.4 & 41.8 & 77.3 & 2.5 & 88.2 & 48.0 & 84.3 & 64.6 & 70.2 & 56.1 & 38.9 \\
			\cmidrule{2-22}
			& PCB-RS & 55.3& 93.8 & 12.4 & 42.3 & 57.7 & 46.4 & 54.6 & 77.3 & 0.0 & 92.2 & 41.3 & 78.9 & 0.2 & 89.2 & 50.6 & 84.6 & 61.2 & 72.5 & 54.4 & 40.3 \\
			\cmidrule{2-22}
			& PCB-RS+SCL & \textbf{56.5}& 94.5 & 6.8 & 35.5 & 76.7 & 37.6 & 48.3 & 79.0 & 0.0 & 93.4 & 42.4 & 80.8 & 1.1 & 89.0 & 56.9 & 87.6 & 68.4 & 74.4 & 61.0 & 40.8 \\
			\toprule[1.5pt]	
		\end{tabular}%
	}
		\vspace{-3mm}
\end{table*}

\begin{table}[ht]
	\centering
	\renewcommand{\arraystretch}{1.3}
	\caption{Evaluation results on the SemanticPOSS dataset. \dag: result reported by~\cite{pan2020semanticposs} \ddag: result reported by~\cite{li2021rethinking} and \cite{li2021multi} *: result reported by our implementation.}
	\resizebox{\linewidth}{!}{%
		\begin{tabular}{r|ccccccccccc|c} 
		\hline
		& person & rider & car & trunk & plants & traffic sign & pole & building & fence & bike & road & mIoU \\ 
		\hline
		PointNet++~\dag\cite{qi2017pointnet} & 20.8 & 0.1 & 8.9 & 4.0 & 51.2 & 21.8 & 3.2 & 42.7 & 6.0 & 0.1 & 62.2 & 20.1 \\
		SqueezeSegV2~\dag\cite{wu2019squeezesegv2} & 18.4 & 11.2 & 34.9 & 15.8 & 56.3 & 11.0 & 4.5 & 47.0 & 25.5 & 32.4 & 71.3 & 29.8 \\ 
		\hline
		RangeNet53~\ddag\cite{milioto2019rangenet++} & 10.0 & 6.2 & 33.4 & 7.3 & 54.2 & 5.5 & 2.6 & 49.9 & 18.4 & 28.6 & 63.5 & 25.4 \\
		RangeNet53 + KNN~\ddag\cite{milioto2019rangenet++} & 14.2 & 8.2 & 35.4 & 9.2 & 58.1 & 6.8 & 2.8 & 55.5 & 28.8 & 32.2 & 66.3 & 28.9 \\
		UnPNet~\ddag\cite{li2021rethinking} &11.3 &12.1 &36.8 &10.6 &62.3 &6.9 &4.2 &60.4 &20.6 &35.4 &65.6 &29.7 \\
		UnPNet + KNN~\ddag\cite{li2021rethinking} &17.7 &17.2 &39.2 &13.8 &67.0 &9.5 &5.8 &66.9 &31.1 &40.5 &68.4 &34.3 \\ 
		MINet~\ddag\cite{li2021multi} & 13.3 & 11.3 & 34.0 & 18.8 & 62.9 & 11.8 & 4.1 & 55.5 & 20.4 & 34.7 & 69.2 & 30.5 \\
		MINet + KNN~\ddag\cite{li2021multi} & 20.1 & 15.1 & 36.0 & 23.4 & 67.4 & 15.5 & 5.1 & 61.6 & 28.2 & 40.2 & 72.9 & 35.1 \\
		\hline
		SalsaNext~*\cite{cortinhal2020salsanext} & 62.6 & 49.8 & 63.5 & 34.7 & 78.0 & 39.1 & 26.7 & 79.4 & 54.7 & 54.2 & 83.1 & 56.9 \\
		SalsaNext + KNN~*\cite{cortinhal2020salsanext} & 62.8 & 50.3 & 64.1 & 35.0 & 78.3 & 39.4 & 27.1 & 80.0 & 54.6 & 54.1 & 82.8 & 57.1 \\
		FIDNet~*\cite{zhao2021fidnet} & 58.1 & 44.5 & 69.3 & 39.9 & 76.6 & 36.4 & 23.6 & 76.5 & 59.6 & 53.4 & 81.7 & 56.3 \\
		FIDNet + KNN~*\cite{zhao2021fidnet} & 58.6 & 45.1 & 70.0 & 40.2 & 76.9 & 37.3 & 23.5 & 77.1 & 59.1 & 53.1 & 81.3 & 56.6 \\ 
		\hline
		BAF-LAC* (Baseline with RS) & 50.4 & 48.7 & \textbf{54.4} & 40.0 & 72.5 & 22.4 & 28.1 & 74.5 & 42.0 & 48.8 & 81.0 & 51.2  \\
		BAF-LAC (Baseline with PCB-RS) & 52.9 & 54.1 & 44.5 & 35.7 & 73.5 & 28.9 & 42.8 & \textbf{78.9} & 52.2 & 49.1 & 82.2 & 54.1  \\
		PCB-BAF-LAC (PCB-RS + SCL) & \textbf{56.4} & \textbf{57.7} & 53.8 & \textbf{43.5} & \textbf{73.9} & \textbf{32.1} & \textbf{43.3} & 72.5 & \textbf{52.8} & \textbf{51.4} & \textbf{83.8} & \textbf{56.5}  \\
		\hline
		BAAF* (Baseline with RS) & 53.4 & 56.1 & 53.8 & 43.5 & 74.3 & 32.7 & 30.7 & 74.7 & 43.4 & \textbf{53.6} & 81.9 & 54.4 \\
		BAAF (Baseline with PCB-RS) & 56.7 & 57.6 & 60.5 & \textbf{45.8} & 77.6 & \textbf{37.0} & 36.5 & 81.4 & 48.7 & 50.4 & 83.7 & 57.8 \\
		PCB-BAAF (PCB-RS + SCL) & \textbf{58.1} & \textbf{57.6} & \textbf{63.2} & 41.8 & \textbf{77.9} & 28.9 & \textbf{33.8} & \textbf{79.8} & \textbf{60.1} & 52.2 & \textbf{84.9} & \textbf{58.0}  \\ 
		\hline	
		RandLA-Net* (Baseline with RS) & 56.8 & \textbf{58.2} & 48.1 & 37.8 & 73.4 & 22.3 & \textbf{35.4} & 76.6 & 45.4 & 45.7 & 80.9 & 52.8 \\
		RandLA-Net (Baseline with PCB-RS) & 55.9 & 53.8 & 58.9 & 44.7 & 77.5 & 34.5 & 27.7 & 82.8 & 42.6 & 49.9 & 82.0 & 55.5 \\
		PCB-RandNet (PCB-RS + SCL) & \textbf{57.7} & 52.4 & \textbf{61.4} & \textbf{42.0} & \textbf{78.1} & \textbf{31.8} & 33.0 & \textbf{81.6} & \textbf{51.5} & \textbf{51.2} & \textbf{84.3} & \textbf{56.8}  \\ 
		\hline
	\end{tabular}}
	\label{tab:result_poss}
	\vspace{-5mm}
\end{table}

\noindent\textbf{Results on SemanticKITTI:} 
Table~\ref{tab:result_kitti} shows the segmentation performance of our proposed method compared with the baseline method for different distance ranges. The experimental results show that our proposed method can improve the segmentation performance of the model at different distance ranges. Specifically, our proposed PCB-RS method brings $1.9\%$, $2.2\%$, and $1.9\%$ mIoU performance improvement over the baseline RS sampling method in the three medium and long distance ranges ($30m \sim 40m$, $40m \sim 50m$, and $\textgreater 50m$), respectively. Combined with the SCL module, the model further achieves performance improvements of $2.3\%$, $2.3\%$, $3.1\%$, $5.5\%$, $4.2\%$, and $4.6\%$ mIoU in each of the six distance interval ranges. 

\noindent\textbf{Results on SemanticPOSS:} Table~\ref{tab:result_poss} shows the comparison of our proposed method with other related works and baseline. To make a fair comparison, we also reproduce and report the results of two more recent and powerful models,  SalsaNext~\cite{cortinhal2020salsanext} and FIDNet~\cite{zhao2021fidnet}, on the SemanticPOSS dataset. Experimental results show that the proposed PCB-RS sampling method and SCL loss actually contribute to significant improvements on more challenging dataset with much smaller objects. 

Specifically, for the baseline approach \textbf{RandLA-Net}, the introductions of PCB-RS and SCL help achieve mIoU improvements of $2.7\%$ and $1.3\%$, respectively, and their combination improves the segmentation performance from $52.8\%$ to $56.8\%$. For \textbf{BAF-LAC}, the usage of PCB-RS and SCL makes improvements of $2.9\%$ and $2.4\%$ mIoU, respectively, and they together improve the performance from $51.2\%$ to $56.5\%$. For \textbf{BAAF}, PCB-RS and SCL improve the mIoU by $2.4\%$ and $1.0\%$, respectively. The segmentation performance obtained by integrating both PCB-RS and SCL is increased from $54.4\%$ to $58.0\%$.

\section{Ablation Study} 
\label{sec:abla}
We conducted the following ablation experiments on the SemanticKITTI dataset to quantify the effectiveness of the different components. To perform efficient training and evaluation, the training and validation sets are built by selecting one frame from the original sequence with every $3$ frames interval. i.e. only $25\%$ of the original training and validation data used.

\subsection{Effects of Polar Cylinder Balanced Random Sampling.}
Table~\ref{ablation:sampling} shows the segmentation performance of our model at different distance ranges with different sampling methods. Considering the randomness of the whole framework, the results under $3$ different random seeds are reported for a fairer comparison. These results show that: 1) The segmentation performance decreases sharply as the distance to the center LiDAR sensor increases. 2) Compared to baseline RS, our proposed PCB-RS sampling method achieves remarkable performance improvements in both medium and long ranges. This is consistent with our expectation that PCB-RS can preserve the points at medium and long distances well. 3) Surprisingly, PCB-RS achieves moderate improvement in the close range ($\textless10m$). We think it is due to the overfitting of the model, and most of the points obtained after RS sampling mainly aggregate in the closer regions, and the network is likely to overfit these closer features, which leads to performance degradation.

\begin{table}[t]
	\centering
	\renewcommand{\arraystretch}{1.4}
	\caption{ Comparison of mean IoU (\%) for different distance ranges and overall on the SemanticKITTI valid set. }
	\resizebox{0.95\linewidth}{!}{
		\begin{tabular}{c|ccc|ccc|ccc} 
			\hline
			& \multicolumn{9}{c}{Seed} \\ 
			\cline{2-10} 
			& \multicolumn{3}{c|}{143} & \multicolumn{3}{c|}{520} & \multicolumn{3}{c}{1024} \\
			& RS & PCB-RS & Improv & RS & PCB-RS & Improv & RS & PCB-RS & Improv \\ 
			\hline
			Range From 0 m to 10 m & 56.67 & 58.56 & +1.89 & 55.71 & 58.24 & +2.53 & 54.85 & 58.63 & +3.78 \\
			Range From 10 m to 20 m & 49.80 & 52.84 & +3.04 & 51.25 & 53.62 & +2.37 & 50.96 & 51.32 & +0.36 \\
			Range From 20 m to 30 m & 40.06 & 43.83 & +3.77 & 39.83 & 44.14 & +4.31 & 38.44 & 42.33 & +3.89 \\
			Range From 30 m to 40 m & 28.94 & 33.06 & +4.12 & 30.10 & 33.28 & +3.18 & 28.36 & 31.62 & +3.26 \\
			Range From 40 m to 50 m & 21.46 & 25.76 & +4.30 & 22.51 & 24.96 & +2.45 & 21.71 & 23.78 & +2.07 \\
			Range Above 50 m & 8.25 & 12.17 & +3.92 & 9.90 & 11.42 & +1.52 & 8.62 & 11.18 & +2.56 \\ 
			\hline
			Total Range & 52.81 & 55.61 & +2.80 & 52.65 & 55.61 & +2.96 & 52.02 & 55.04 & +3.02 \\
			\hline
	\end{tabular}}
	\label{ablation:sampling}
\end{table}

\begin{table}[!t]
	\centering
	\caption{Effect of of Resolution of Polar Cylinder on SemanticKITTI valid set.}
	\label{resolution}
		\resizebox{\linewidth}{!}{%
		\begin{tabular}{c|ccc} 
			\toprule
			Resolution of Polar Cylinder~ & None (Baseline with RS) & 32$\times$32$\times$16 & 48$\times$48$\times$16 \\ 
			\midrule
			mIoU(\%) & 52.02 & 54.34 & 54.67 \\ 
			\toprule
			Resolution of Polar Cylinder~ & None (Baseline with RS) & 64$\times$64$\times$16 & 96$\times$96$\times$32 \\ 
			\midrule
			mIoU(\%) & 52.02 & \textbf{55.04} & 53.46 \\
			\toprule
		\end{tabular}
		}
	\vspace{-7mm}
\end{table}

\subsection{Effect of Resolution of Polar Cylinder.}
Ablation studies about resolution of the polar cylinder representation are reported in Table~\ref{resolution}. The results show that PCB-RS is relatively robust to the size setting of the resolution of polar cylinder.

\subsection{Effects of Sampling Consistency Loss.}
Here, we conduct further experiments to validate the effectiveness of the proposed sampling consistency loss and uncertainty weighting methods. As shown in Table~\ref{ablation:scl}, fixed weighting weights only lead to slight performance improvements or even performance degradation, while the learnable weighting method shown in Equation~\ref{eq:learn} can bring significantly better performance. Moreover, the validation results on the sub-sample set show that the introduction of sampling consistency loss can reduce the performance variance of the model under different sampling methods, and the learnable weighting method can reduce this variance (from $1.99\%$ miou to $1.07\%$ miou).

\subsection{Consistent Effectiveness under Different Baselines.}
We have verified the effectiveness of PCB-RS and SCL on three different baseline networks (RandLA-Net, BAF-LAC and BAAF) on the SemanticPOSS dataset in Section~\ref{sec:result}. However, considering the obvious domain differences or sensor differences between the SemanticKITTI and SemanticPOSS datasets, it is necessary to perform validation on the SemanticKITTI dataset. In order to avoid the excessive time and resource consumption required for training with the whole data, we only perform fast validation on a smaller sub-sample dataset in the ablation experiment. The experimental results in Table~\ref{ablation:backbone} show that our PCB-RS sampling method and SCL loss make the BAF-LAC and BAAF achieve consistent enhancement results. Meanwhile, as we expected, the improvement of the model at medium and long distances is more obvious.

\begin{table}[!t]
	\renewcommand{\arraystretch}{1.2}
	\centering
	\caption{Evaluation of sampling consistency loss and uncertainty weighting method.}
	\label{ablation:scl}
		\begin{tabular}{c|c|ccc} 
			\hline
			\multicolumn{1}{l|}{\multirow{2}{*}{Loss Function}} & \multicolumn{1}{l|}{mIoU on val set} & \multicolumn{3}{l}{mIoU on subsample set} \\ 
			\cline{2-5}
			\multicolumn{1}{l|}{} & PCB-RS & RS & PCB-RS & Gap \\ 
			\hline
			$\mathcal{L}_{wce}$ & 55.04 & 48.48 & 50.47 & 1.99 \\
			$\mathcal{L}_{wce} + 10*\mathcal{L}_{scl}$ & 55.34 & 50.14 & 51.44 & 1.30 \\
			$\mathcal{L}_{wce} + 15*\mathcal{L}_{scl}$ & 54.69 & 51.01 & 52.39 & 1.38 \\
			Equation~\ref{eq:learn} & 57.16 & 53.07 & 54.14 & 1.07 \\
			\hline
		\end{tabular}
\end{table}

\section{Limitation and Future Work.}
Since we only study the effect of the sampling method in this paper, we do not make any adjustments to the baseline network structure~\cite{hu2020randla, shuai2021backward, qiu2021semantic}. However, the point cloud distributions under RS and PCB-RS sampling methods are very different. Therefore it is necessary to design more suitable local feature extraction and local feature aggregation operators for PCB-RS sampling. This will be considered as our future work.

\begin{table}[t]
	\renewcommand{\arraystretch}{1.2}
	\centering
	\caption{Evaluation of PCB-RS and SCL on different baselines.}
	\label{ablation:backbone}
	\resizebox{\linewidth}{!}{
		\begin{tabular}{r|c|cc|cc} 
			\toprule[1.5pt]
			& \begin{tabular}[c]{@{}c@{}}BAF-LAC\\(Baseline with RS)\end{tabular} & \begin{tabular}[c]{@{}c@{}}BAF-LAC\\(PCB-RS)\end{tabular} & Improv & \begin{tabular}[c]{@{}c@{}}PCB-BAF-LAC\\(PCB-RS+SCL)\end{tabular} & Improv \\ 
			\midrule
			Range From 0 m to 10 m & 56.44 & 57.50 & +1.06 & 56.96 & +0.52 \\
			Range From 10 m to 20 m & 49.74 & 51.84 & +2.10 & 53.44 & +3.70 \\
			Range From 20 m to 30 m & 39.23 & 42.05 & +2.82 & 44.98 & +5.75 \\
			Range From 30 m to 40 m & 28.71 & 32.89 & +4.15 & 36.30 & +7.59 \\
			Range From 40 m to 50 m & 22.89 & 25.02 & +2.13 & 27.03 & +4.14 \\
			Range Above 50 m & 9.80 & 11.11 & +1.31 & 13.10 & +3.30 \\
			\midrule
			Total Range & 52.96 & 54.65 & +1.69 & 55.16 & +2.20 \\
			\toprule[1.5pt]
			& \begin{tabular}[c]{@{}c@{}}BAAF\\(Baseline with RS)\end{tabular} & \begin{tabular}[c]{@{}c@{}}BAAF\\(PCB-RS)\end{tabular} & Improv & \begin{tabular}[c]{@{}c@{}}PCB-BAAF\\(PCB-RS+SCL)\end{tabular} & Improv \\ 
			\midrule
			Range From 0 m to 10 m & 55.47 & 57.71 & +2.24 & 57.96 & +2.49 \\
			Range From 10 m to 20 m & 50.74 & 51.25 & +0.51 & 52.68 & +1.94 \\
			Range From 20 m to 30 m & 40.32 & 42.59 & +2.27 & 45.24 & +4.92 \\
			Range From 30 m to 40 m & 29.37 & 31.23 & +1.86 & 34.30 & +4.93 \\
			Range From 40 m to 50 m & 21.57 & 22.97 & +1.40 & 24.78 & +3.21 \\
			Range Above 50 m & 10.49 & 9.84 & -0.65 & 13.13 & +2.64 \\
			\midrule
			Total Range & 52.53 & 54.20 & +1.67 & 55.58 & +3.05 \\
			\toprule[1.5pt]
	\end{tabular}}
	\vspace{-5mm}
\end{table}

\section{Conclusion}
In this paper, we propose a novel sampling method, called Polar Cylinder Balanced Random Sampling, for the autonomous driving LiDAR point cloud segmentation task. The PCB-RS sampling method enables the sampled point clouds to maintain a more balanced distribution and guides the network to fully learn the point cloud features at different distances. We also propose a sampling consistency loss to further improve the model performance and reduce the variance of the model under different sampling methods. Experimental results on SemanticKITTI and SemanticPOSS show that our proposed approach achieves consistent performance improvements and enhances the segmentation performance of the model at different distance ranges.

\clearpage
\bibliographystyle{IEEEtran}
\bibliography{IEEEexample}

\end{document}